\documentclass[conference]{IEEEtran}
\IEEEoverridecommandlockouts
\usepackage{cite}
\usepackage{amsmath,amssymb,amsfonts}
\usepackage{algorithmic}
\usepackage{graphicx}
\usepackage{textcomp}
\usepackage{xcolor}
\usepackage{multirow}
\usepackage{hyperref}
\usepackage{orcidlink}

\makeatletter 
\newcommand{\linebreakand}{%
  \end{@IEEEauthorhalign}
  \hfill\mbox{}\par
  \mbox{}\hfill\begin{@IEEEauthorhalign}
}
\makeatother 

\newcommand\independent{\protect\mathpalette{\protect\independenT}{\perp}}
\def\independenT#1#2{\mathrel{\rlap{$#1#2$}\mkern2mu{#1#2}}}
\def\BibTeX{{\rm B\kern-.05em{\sc i\kern-.025em b}\kern-.08em
    T\kern-.1667em\lower.7ex\hbox{E}\kern-.125emX}}

\author{
    \IEEEauthorblockN{
        Omid Orang$^{\orcidlink{0000-0002-4077-3775}}$\IEEEauthorrefmark{5},
        Patrícia O. Lucas$^{\orcidlink{0000-0002-7334-8863}}$\IEEEauthorrefmark{2}\IEEEauthorrefmark{3}\IEEEauthorrefmark{5},
        Gabriel I. F. Paiva\IEEEauthorrefmark{4}\IEEEauthorrefmark{5},
        Petrônio C. L. Silva$^{\orcidlink{0000-0002-1202-2552}}$\IEEEauthorrefmark{1}\IEEEauthorrefmark{5},\\
        Felipe Augusto Rocha da Silva$^{\orcidlink{0000-0003-4567-8504}}$\IEEEauthorrefmark{3}\IEEEauthorrefmark{5},
        Adriano Alonso Veloso$^{\orcidlink{0000-0002-9177-4954}}$\IEEEauthorrefmark{5}
        and Frederico Gadelha Guimarães$^{\orcidlink{0000-0001-9238-8839}}$\IEEEauthorrefmark{5}}
 \IEEEauthorblockA{
 \IEEEauthorrefmark{1}
 \textit{Federal Institute of Northern Minas Gerais}, Januária, MG, Brazil  \\ 
 \IEEEauthorrefmark{2}
 \textit{Federal Institute of Northern Minas Gerais}, Salinas, MG, Brazil  \\ 
 \IEEEauthorrefmark{3}
 \textit{Graduate Program in Electrical Engineering}, \textit{Universidade Federal de Minas Gerais}, \\
 Av. Antônio Carlos 6627, 31270-901, Belo Horizonte, MG, Brazil  \\  
 \IEEEauthorrefmark{4}
 \textit{Graduate Program in Computer Science}, \textit{Federal University of Minas Gerais}, Belo Horizonte, Brazil  \\  
 \IEEEauthorrefmark{5}
 \textit{Future Lab},
 \textit{Department of Computer Science},
 \textit{Federal University of Minas Gerais},
 Belo Horizonte, Brazil\\ Email: omid.orang@dcc.ufmg.br, \{patricia.lucas, petronio.candido\}@ifnmg.edu.br, \\ gabrielpaiva18@gmail.com, fars@ufmg.br, adrianov@dcc.ufmg.br and fredericoguimaraes@ufmg.br
 }
 }

\begin{document}

\title{Causal Graph Fuzzy LLMs: A First Introduction and Applications in Time Series Forecasting}


\maketitle

\begin{abstract}
In recent years, the application of Large Language Models (LLMs) to time series forecasting (TSF) has garnered significant attention among researchers. This study presents a new frame of LLMs named CGF-LLM using GPT-2 combined with fuzzy time series (FTS) and causal graph to predict multivariate time series, marking the first such architecture in the literature. The key objective is to convert numerical time series into interpretable forms through the parallel application of fuzzification and causal analysis, enabling both semantic understanding and structural insight as input for the pretrained GPT-2 model.  The resulting textual representation offers a more interpretable view of the complex dynamics underlying the original time series. The reported results confirm the effectiveness of our proposed LLM-based time series forecasting model, as demonstrated across four different multivariate time series datasets. This initiative paves promising future directions in the domain of TSF using LLMs based on FTS.

\end{abstract}


\section{Introduction}
Time series forecasting (TSF) and analysis play a pivotal role in many real-world applications, such as energy, traffic, healthcare, finance, and meteorology \cite{Abdullahi2025,liu2025timeseriesanalysisbenefit}. A variety of forecasting methods have been proposed in the literature, generally classified into three main categories: statistical, machine learning (ML), and deep learning (DL) approaches \cite{kim2025comprehensivesurveydeeplearning}. Despite their successes, these methods still face challenges such as high dimensionality, missing values, limited data availability, and the need to capture long-term dependencies—all of which are critical for accurate temporal modeling \cite{Abdullahi2025}.

Large Language Models (LLMs), built on transformers composed of billions of parameters, have emerged to revolutionize DL models \cite{kim2025comprehensivesurveydeeplearning}. Although LLMs were originally developed for natural language processing (NLP) tasks, their application in time series (TS) analysis and forecasting has recently gained significant momentum. This shift is largely due to their ability to leverage self-attention mechanisms to capture temporal dependencies and complex dynamics, thereby effectively modeling input–output relationships \cite{chow2024timeseriesreasoningllms}. This surge is driven by their remarkable ability to capture long-range dependencies and complex sequential patterns through the attention mechanism inherent in the transformers \cite{Abdullahi2025,ekambaram2024tinytimemixersttms}. 

According to the reviewed literature \cite{jiang2024empowering,Abdullahi2025}, a number of methods leveraging LLMs have been proposed for TSF, differing in input types, model integration strategies, and LLM architectures. For instance, Time-LLM \cite{jin2023time} uses LLaMA and GPT-2 to process multimodal TS (MTS) and relies on tokenization and fine-tuning strategies. TEMPO \cite{cao2023tempo}, which also employs GPT-2, handles univariate TS data using tokenization and prompt-based modeling without fine-tuning. In contrast, PromptCast \cite{Xue2024promptcast} uses BART and BERT, focusing on prompt engineering without modifying LLM weights. Chronos \cite{ansari2024chronoslearninglanguagetime} incorporates GPT-2 and T5, combining prompt design with model fine-tuning. LLMTIME \cite{gruver2024largelanguagemodelszeroshot} utilizes GPT-3 and LLaMA-2 to handle multimodal inputs with prompt tuning and partial integration of LLMs into the downstream task. GPT4MTS \cite{Jia2024GPT4MTS}, employing GPT-2, and UniTime \cite{Liu2024Unitime}, also with GPT-2, both use prompt-based methods without integrating the LLM as part of the model. Meanwhile, S\textsuperscript{2}IP-LLM \cite{Pan2024S2IP-LLM} relies on GPT-2 and fully integrates it into the forecasting pipeline. Several methods, such as LAMP   \cite{shi2023language} (using GPT-3 variants and LLaMA-2) and the model proposed in  \cite{yu2023temporal} (with GPT-4 and Open-LLaMA), explore newer LLMs and fine-tuning for multivariate forecasting in specific domains. These diverse strategies reflect the flexibility of LLMs in modeling sequential dependencies in TS, originally designed for text but increasingly adapted to structured temporal data. Notably, methods like Time-LLM, TEMPO, Chronos, and S\textsuperscript{2}IP-LLM provide open-source code, fostering reproducibility and further research.

In more domain-specific scenarios, the authors in \cite{lopezlira2024chatgptforecaststockprice} use ChatGPT to query multimodal data for financial forecasting without fine-tuning or integration. In the healthcare domain, Liu et al. \cite{liu2023large} apply PaLM for MTS both forecasting and classification, although without integration or tokenization. For mobility forecasting, AuxMobLCast \cite{xue2022leveraging} leverages LLMs such as BERT, RoBERTa, GPT-2, and XLNet, combining tokenization and fine-tuning strategies. LLM-Mob \cite{wang2024inextlargelanguage} builds on GPT-3.5, using token-based modeling without integration. Finally, in traffic forecasting, ST-LLM \cite{liu2024spatial} employs LLaMA and GPT-2, utilizing tokenization and prompt engineering with full integration into the final model. These models demonstrate that LLMs can be effectively adapted beyond general domains, extending their capabilities to temporal, multimodal, and spatio-temporal forecasting tasks across sectors. 

This research pioneers a new multiple-input single-output (MISO) LLM-based forecasting model termed CGF-LLM. This method combines the concepts of fuzzy time series (FTS) \cite{lucas2022tutorial}, causal graphs, and LLMs. The central objective of this work is to transform numerical time series into interpretable linguistic representations through the parallel application of fuzzification and causal analysis. This dual approach enables both semantic understanding of variable behavior and structural insight into their temporal dependencies. Specifically, the framework integrates FTS modeling with causal discovery using the PCMCI algorithm \cite{PCMCI_2019}. By combining these two perspectives, the method constructs a fuzzy causal text that is both data-driven and interpretable, which serves as input for GPT-2. In other words, it extracts meaningful knowledge, providing a clearer understanding of the complex dynamics within the original time series and revealing the causal relationships among variables. The results confirm that the proposed CGF-LLM technique surpasses the standard LLM in both accuracy and computational efficiency.

The remainder of this paper is structured as follows: Section~\ref{sec:Preliminary_Concepts} provides the basics of LLMs, FTS, and causal graphs. Section \ref{sec_CGF} outlines the details of the proposed CGF-LLM method. Section \ref{sec_experiments} covers the case studies, results, and discussion. Finally, Section \ref{sec_conclusao} concludes the paper and highlights the future research avenues.  

\section{Preliminary Concepts}
\label{sec:Preliminary_Concepts}

\subsection{Large Language Models}
\label{sec_LLMs}
The advent of Large Language Models (LLMs) has significantly advanced natural language processing (NLP), with early landmark models such as BERT \cite{devlin2019bert}, GPT-2 \cite{radford2019language}, and RoBERTa \cite{liu2019roberta}. These models are built upon the Transformer architecture introduced in \cite{attentionvaVaswani2017}, which replaced recurrence with self-attention, enabling more effective handling of sequential data.

A major leap occurred with GPT-3 \cite{Brown2023}, a 175 B-parameter model that demonstrated remarkable few-shot capabilities across diverse tasks. Its success, however, also highlighted issues such as limited transparency, bias, and occasional unreliability. Subsequent models sought to improve on these limitations. Google’s PaLM \cite{Palm2022} introduced the Pathways approach for better reasoning and coherence at scale (540B parameters), while DeepMind’s Chinchilla \cite{hoffmann2022training} showed that smaller models (70B) can outperform larger ones when trained more efficiently. Recent models such as GPT-4 \cite{openai2024gpt4technicalreport}, LaMDA \cite{thoppilan2022lamdalanguagemodelsdialog}, LLaMA \cite{touvron2023llama2023}, and DeepSeek 
\cite{deepseek-R12025}, among others, further expand capabilities. 

Alongside performance gains, efforts to enhance efficiency (e.g., pruning, quantization) and interpretability (e.g., attention maps, chain-of-thought prompting) have become central to LLM research, enabling broader and safer application across domains, including time series analysis.

\subsection{Fuzzy time series}
\label{sec_FTS}

The concept of Fuzzy Time Series (FTS) was originally proposed by \cite{song1993forecasting,song1994forecasting,chen1996forecasting}, based on the fuzzy set theory of \cite{zadeh1965fuzzy}. The main idea of the model is to convert numerical time series into linguistic representations to describe and forecast their behavior through fuzzy relation rules or matrices.

Consider a univariate time series $Y \in \mathbb{R}$, composed of observations $y(t)$ for $t = 0, 1, \dots, T$. Each value $y(t)$ can be associated with a fuzzy set $A_i \in \tilde{A}$ through a membership function $\mu_{A_i}: \mathbb{R} \rightarrow [0, 1]$, which measures the degree of membership of $y(t)$ in $A_i$. The most common membership functions are triangular, trapezoidal, sigmoid, and Gaussian.

According to the approach of \cite{chen1996forecasting}, the training process of an FTS model is divided into three main steps:

\begin{enumerate}
\item[a)] \textit{Partitioning}: Universe of Discourse (UoD), represented by $U = [\min(Y), \max(Y)]$, is segmented into $k$ overlapping subintervals. For each subinterval, a fuzzy set $A_i$ is defined with its respective membership function $\mu_{A_i}$ and central point $c_i$. From these sets, the linguistic variable $\tilde{A}$ is constructed, whose terms are the $A_i$, with $i = 1, \dots, k$.

\item[b)] \textit{Fuzzification}: transforms the series $Y$ into a fuzzy sequence $F$, in which each element $f(t) \in F$ corresponds to a $k$-component vector, representing the degrees of association of the value $y(t)$ with the fuzzy sets $A_i$ of the linguistic variable $\tilde{A}$.

\item[c)] \textit{Rule Generation}: from the fuzzified series $F$, transition patterns are identified between consecutive pairs $(f(t-1), f(t))$, from which rules of the type $A_i \rightarrow A_j, A_k, \dots$ are extracted. These rules represent relationships such as: “IF $f(t)$ is $A_i$, THEN $f(t+1)$ is $A_j$, $A_k$, etc.”.
\end{enumerate}

The FTS model, therefore, is composed of both the linguistic variable $\tilde{A}$ and the extracted set of fuzzy rules. To perform forecasting, the model follows these steps:

\begin{enumerate}
\item[a)] \textit{Input Fuzzification}: the current value $y(t)$ is converted into its fuzzy representation $f(t)$ using the previously defined membership functions.

\item[b)] \textit{Rule Activation}: the subset of rules $R$ is identified whose antecedents contain the fuzzy set corresponding to $f(t)$. The activation degree $\mu_r$ of each rule $r \in R$ is given by the corresponding membership value.

\item[c)] \textit{Defuzzification}: the predicted value $y(t+1)$ is calculated based on a weighted average of the central points $mp_r$ of the activated rules. The central point of a rule is given by:
\begin{equation}
mp_r = \sum_{i \in consequent} c_i
\end{equation}
The final forecast is then obtained by:
\begin{equation}
y(t+1) = \frac{\sum_{r \in R} \mu_r \cdot mp_r}{\sum_{r \in R} \mu_r}
\end{equation}
\end{enumerate}

\subsection{Causal discovery in multivariate time series}
\label{sec_PCMCI}

In this part the PCMCI method developed by \cite{PCMCI_2019} is discussed. The PCMCI is a causal discovery method that generates a causal graph directly from multivariate time series. It consists of two steps: first, using the PC$_1$ algorithm to identify the parents $\hat{\mathcal{P}}(X_t^j)$ for all variables in the time series $X_t^j \in {X_t^1, \cdots ,X_t^N}$, and second, applying the momentary conditional independence (MCI) test to test for indirect links $X_{t-\tau}^i \longrightarrow X_t^j$.

PC$_1$ is an algorithm that uses iterative independence tests for discovering Markov sets. For each variable $X^j_t$, the initialization of the initial parents $\mathcal{\hat{P}}(X_t^j) = (X_{t-1},X_{t-2}, \cdots, X_{t-\tau_{max}})$ is performed. First, unconditional independence tests are applied to remove $X^i_{t-\tau}$ from $\mathcal{\hat{P}}(X_t^j)$ if the null hypothesis $X^i_{t-\tau} \independent X_t^j$ is not rejected given a certain significance level $\alpha_{PC}$. The initial parents are ranked by their absolute test statistic value. 

\begin{figure}[h]
\centerline{\includegraphics[scale=0.15]{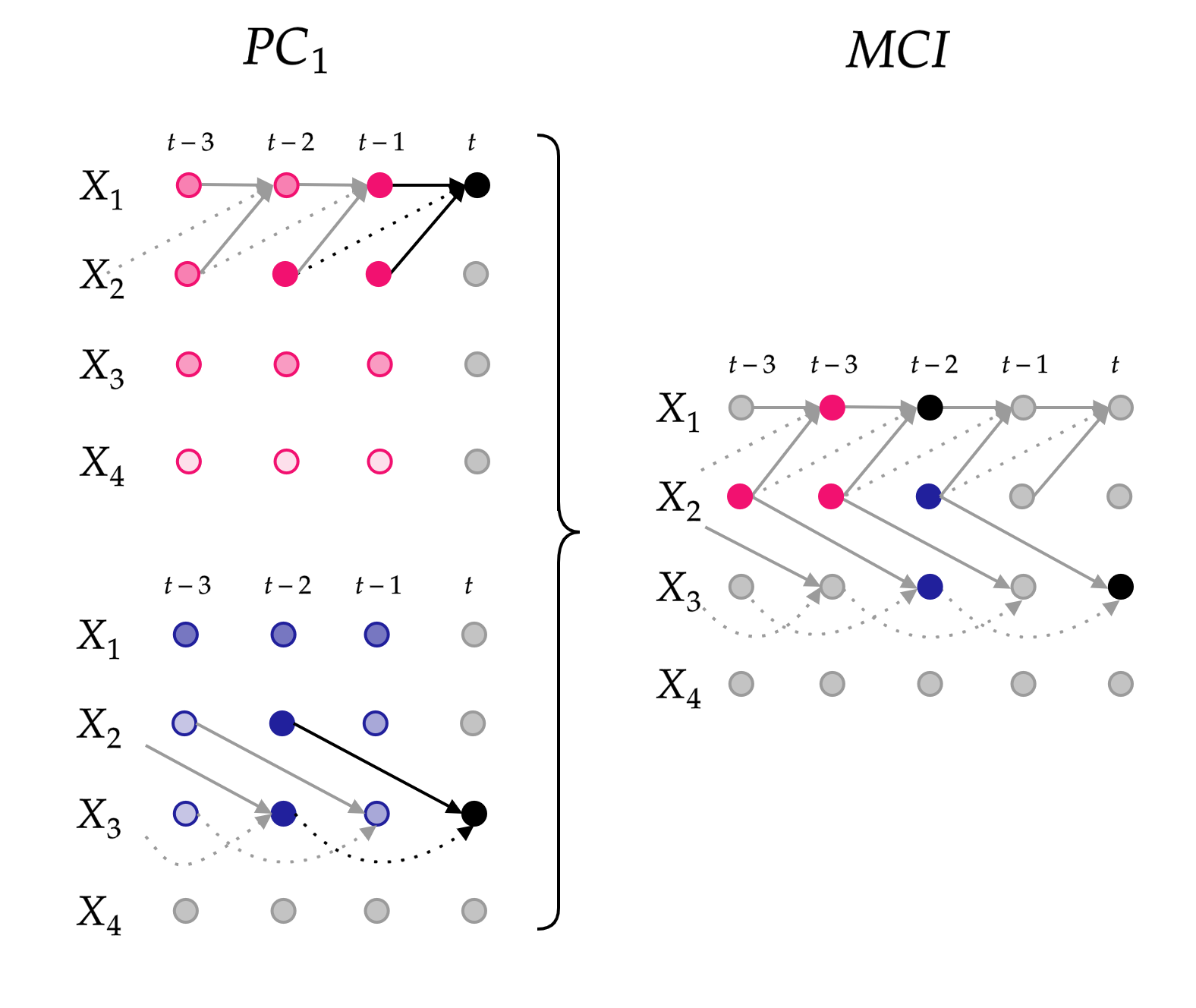}}
\caption{Illustration of the PCMCI method.}
\label{fig:pcmci}
\end{figure}

Figure \ref{fig:pcmci} illustrates PC$1$ for two variables $X_1$ and $X_3$ where the color intensity represents the absolute test statistic value of the dependent variables (darker color indicates higher value). Gray nodes represent independent variables. Then, conditional independence tests $X^i_{t-\tau} \independent X^j_t | \mathcal{L}$ are performed, where $\mathcal{L}$ represents the strongest parents in $\mathcal{\hat{P}}(X_t^j) \smallsetminus {X^i_{t-\tau}}$, and the independent parents are removed from $\mathcal{\hat{P}}(X_t^j)$. In this way, PC$_1$ converges to only a few relevant conditions that include causally linked parents with high probability (dark pink/dark blue) and potentially some false positives (dashed arrows).

In step (2), the MCI test uses the parents estimated by PC$_1$ to identify indirect causes. In the example illustrated in Figure \ref{fig:pcmci}, the conditions $\mathcal{\hat{P}}(X_t^3)$ are sufficient to establish conditional independence and test $X^1_{t-2} \longrightarrow X^3_{t}$. Additionally, lagged parents from $\mathcal{\hat{P}}(X^1_{t-2})$ are included as additional conditions, and they are responsible for maintaining the false positive rate at an expected level.

The PCMCI method is based on the assumptions of causal sufficiency\footnote{Implying that all common drivers are among the observed variables.}, the Causal Markov Condition, and the faithfulness assumption. It also does not assume contemporaneous causal effects and assumes stationarity \cite{PCMCI_2019}. The PCMCI has a polynomial complexity in the number of variables $N$ and $\tau_{max}$. In the worst-case scenario, where the graph is fully connected, the computational complexity of the PC$1$ condition selection stage for $N$ variables equates to $N^3 \tau{max}^2$. The MCI step involves additional tests of $N^2 \tau_{max}$ (for $\tau > 0$). Therefore, the total worst-case computational complexity in terms of the number of variables is polynomial and given by $N^3 \tau_{max}^2 + N^2 \tau_{max}$.

\section{Causal Graph Fuzzy LLM}
\label{sec_CGF}

This section introduces CGF-LLM, a novel multivariate time series forecasting framework that integrates FTS, causal graphs, and LLMs. The method begins by converting numerical time series into linguistic variables through fuzzification. Next, a causal graph is constructed using the PCMCI algorithm~\cite{PCMCI_2019} to capture temporal and causal relationships among variables. Specifically, the framework integrates FTS modeling with causal discovery using the PCMCI algorithm, where PCMCI identifies which variables are causally relevant, and fuzzification captures how those variables evolve over time. The resulting fuzzy linguistic descriptions and causal relations are combined into interpretable textual representations, which are provided as input to a pretrained GPT-2 model. The LLM generates a linguistic forecast, which is then defuzzified to obtain the final numerical prediction.

\begin{figure*}[!htb]
    \centering
    \includegraphics[scale=0.13]{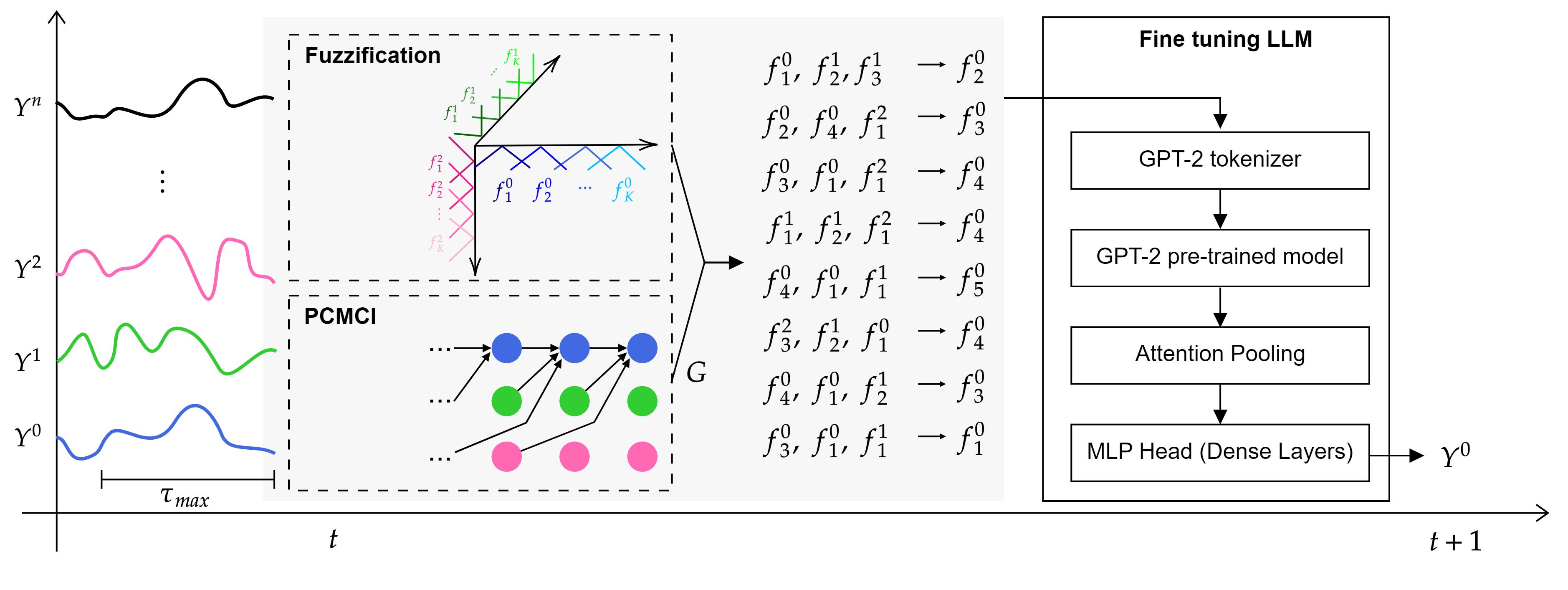}
    \caption{Schematic overview of the proposed CGF-LLM forecasting model. $Y^j$ denotes the input multivariate time series, where $j = 0, \dots, n$. $Y^0$ corresponds to the endogenous time series.  $\tau_{max}$ represents the size of the lag window observed by the PCMCI method.  The fuzzification transforms each $Y^j$ into a FTS $f^j$ using $K$ fuzzy sets. Then, the temporal linguistic patterns are generated using parallel application of fuzzification and PCMCI. For example, the causal relation $f^0_1, f^1_2, f^1_3 \rightarrow f^0_2$ illustrates that the fuzzy representation $f^0_1$ (value of $Y^0$  in fuzzy set 1), $f^1_2$ and $f^1_3$ (values of $Y^1$ in fuzzy sets 2 and 3) influence $f^0_2$, as identified by PCMCI. Arrows in the causal graph ($G$) denote directed causal relationships among fuzzy sets, such that $f^0_1$ to $f^0_2$, $f^1_2$ to $f^0_2$ and others.
    }
    \label{fig:cgf}
\end{figure*}

As illustrated in Figure~\ref{fig:cgf}, CGF-LLM frames the forecasting task as a MISO (Multiple Input Single Output) system, where the $n$ time series variables are treated as inputs $(Y^0, Y^1, Y^2, \dots, Y^n)$, and the output corresponds to the endogenous variable $Y^0$, which is to be predicted at time step $t+1$.

The CGF-LLM methodology comprises four main steps:

\begin{enumerate}
    \item[1)] \textit{\textbf{Fuzzification:}} The Universe of Discourse of the $n$ time series is divided into $K$ overlapping intervals, with each fuzzy set $f^j_i$ defined by its membership function $\mu_{f^j_i}$ and central point $c_{f^j_i}$, where $i = 1, \dots, K$ and $j = 0, \dots, n$. Each time series is then converted into a fuzzy time series $F^j$, where each element $f^j(t) \in F^j$ is a $K$-dimensional vector representing the degrees of membership of the value $y^j(t)$ in the fuzzy sets $f^j_i$.
    
    \item[2)] \textit{\textbf{Causal Graph Construction:}}
    The PCMCI algorithm is used to identify causal relationships among variables, resulting in a causal graph $G$ for the endogenous variable $Y^0$. This graph is constructed based on all input variables, considering a maximum lag window of size $\tau_{\text{max}}$.
    
   \item[3)] \textit{\textbf{Text Generation:}}
   Based on the fuzzy series $f^j$ and the causal graph $G$, temporal linguistic patterns are extracted that reflect the identified dependencies. For example, if $G$ contains the links $Y^0(t-1) \rightarrow Y^0(t)$ and $Y^1(t-1) \rightarrow Y^0(t)$, the resulting pattern will be: $F(Y^0(t-1)), F(Y^1(t-1)) \rightarrow F(Y^0(t))$.
   
    \item[4)] \textit{\textbf{Forecasting:}} The linguistic patterns derived from the FTS and causal graph are used to fine tune the GPT-2 model, originally designed for text generation, for numerical TSF. In more details, pretrained GPT-2 is integrated with additional layers that convert its textual output into a numerical value. This conversion involves a self-attention-based pooling step to aggregate token embeddings into a single representative vector that captures the global context of the sequence. This vector is then processed through a series of linear layers to produce the final output: the numerical forecast of variable $Y^0$ at time $t+1$.
\end{enumerate}

\section{Computational Experiments}
\label{sec_experiments}

\subsection{Case studies}

The datasets were named according to their respective application domains: economic, energy, Internet of Things (IoT), and climate. Table \ref{tab:datasets} provides a summary of the main details of the datasets. Accordingly, the proposed method is tested on these datasets to predict solar radiation in Brazil, wind power generation, household electricity consumption in Mexico, and Bitcoin prices in the United States.

\begin{table*}[!htb]
    \centering
    \caption{Summary of Datasets Used in the Experiments}
    \begin{tabular}{|c|c|c|c|c|c|}
        \hline
        \textbf{Dataset} & \textbf{Description} & \textbf{Target Variable} & \textbf{Samples} & \textbf{Variables} & \textbf{Frequency} \\ \hline
        ECONOMIC & Bitcoin Daily Price Index & AVG & 2.970 & 6 & Daily \\ \hline
        ENERGY & Wind Power Production & Power & 43.800 & 9 & Hourly \\ \hline
        IoT & Household Electricity & active\_power & 100.000 & 14 & Minutely \\ \hline
        CLIMATIC & Solar Radiation & glo\_avg & 35.000 & 12 & Minutely \\ \hline
    \end{tabular}
    \label{tab:datasets}
\end{table*}

\subsection{Experimental Methodology}

The experiments were conducted as an ablation study to assess the impact of the proposed technique (CGF-LLM) on the performance of the GPT-2 model in multivariate time series forecasting. To this end, we compared the methods in terms of prediction accuracy and the number of tokens generated. Three configurations were evaluated: (i) the complete proposed method (CGF-LLM), (ii) a variation without the fuzzyfication step (CG-LLM), and (iii) a baseline LLM, in which numerical time series values were simply converted into text.

Each configuration was trained using two strategies: with freezing, where GPT-2's internal parameters were frozen and only an additional output layer was trained, and no freezing, where both GPT-2 and the additional layers were fine-tuned.

CGF-LLM was configured with the following hyperparameters: lag window size $\tau_{max} = 20$, $\alpha_{PC} = 0.1$, number of partitions set to 30, grid-based partitioning method, and triangular membership function. These values were empirically defined, taking into account the computational cost of running the experiments. The CG-LLM and LLM approaches used the same hyperparameters, adjusted according to their respective architectures. In all three approaches, the GPT-2 model was fine-tuned for 20 epochs.

For the comparative analysis, four datasets from distinct domains and with varying dimensionalities were employed. The experiments were conducted over $10$ time windows, each with a size of $0.3 \cdot |D|$ and an overlap of $30\%$ along the multivariate time series of length $|D|$. Each window was split into training and test subsets, with $20\%$ of the data reserved for testing.

Prediction accuracy was assessed using the average Normalized Root Mean Square Error (NRMSE) computed across the ten windows, as defined in Equation~\eqref{eqn:rmse}. Here, $y_{\max}$ and $y_{\min}$ denote the maximum and minimum values within the test set, respectively.

\begin{equation}
\text{NRMSE} = \frac{\sqrt{\sum_{t=0}^n (y(t) - \hat{y}(t))^2}}{y_{\max} - y_{\min}}
\label{eqn:rmse}
\end{equation}

It is worth noting that all the experiments are implemented and tested in Python 3, using open source libraries including PyFTS \cite{pyfts_2019}, Tigramite\footnote{\href{https://jakobrunge.github.io/tigramite/}{https://jakobrunge.github.io/tigramite/}} and Hugging Face\footnote{\href{https://huggingface.co/docs/transformers/model_doc/gpt2}{https://huggingface.co/docs/transformers/model\_doc/gpt2}}

To promote transparency and reproducibility of the proposed model, source code and datasets are available at https://github.com.

\subsection{Results and Discussion}

Table \ref{tab:nrmse} summarizes the means and standard deviations of the NRMSE for one-step-ahead forecasts across all datasets using three models: CGF-LLM, CG-LLM, and LLM. Two training strategies are compared:
no freezing, where all the parameters of the GPT-2 model are fine-tuned for the forecasting task, and freezing, where only the last two layers (including attention pooling and MLP head) are fine-tuned to reduce the computational demands.

Across all datasets, the no-freezing strategy consistently achieves lower values of average NRMSE compared to those with freezing, indicating that full fine-tuning of the GPT-2 model improves the forecasting accuracy. For example, in the Energy dataset, CGF-LLM records an NRMSE of 0.066 $\pm$ 0.006 without freezing versus 0.153 $\pm$ 0.009 with freezing, highlighting a substantial performance gap. This trend holds across the other three datasets as well. The superior performance of the no-freezing strategy likely stems from its potential to fully adjust the model to the dynamic patterns of datasets, though it needs greater computational resources. 

Comparing the models, CGF-LLM outperforms CG-LLM and LLM in all scenarios, with or without freezing. The exception occurs in the CLIMATIC with freezing, where CG-LLM reports the lowest forecasting error in comparison to CGF-LLM and LLM. In the ECONOMIC dataset, performance differences among the models are minimal, with overlapping standard deviations (e.g., 0.094 $\pm$ 0.018 for CGF-LLM vs. 0.104 $\pm$ 0.018 for LLM without freezing), suggesting no statistically significant distinction.

Thus, the obtained results underscore the efficacy of the proposed CGF-LLM for MISO time series forecasting, particularly under the no-freezing strategy. CG-LLM ranks second in most cases, while LLM generally performs the least effectively. These findings highlight a trade-off between computational efficiency and forecasting precision, with CGF-LLM offering the most robust solution when resources permit fine-tuning the full model. 

In addition to the accuracy enhancement, Table \ref{tab:tokens} reveals the computational efficiency of the proposed CGF-LMM method regarding the number of tokens. Although the total text size produced by the combined use of causal graphs and FTS is greater than that of CG-LLM without fuzzy logic, CGF-LLM results in significantly fewer tokens compared to both CG-LLM and standard LLM approaches. The reason is that fuzzy labels tokenize into fewer units with the GPT-2 tokenizer due to their repetitive nature and simpler structure compared to numerical strings. For instance, a label like “$f_1$” often results in a single token, whereas “23.5” may split into multiple tokens (e.g., “23” and “.5”). Thus, CGF-LLM substantially reduces the number of tokens. For instance, on the IoT dataset, CGF-LLM produces 36 times fewer tokens than the standard LLM (839,412 vs. 30,699,520), leading to a drastic reduction in both memory usage and computational requirements.


In summary, CGF-LLM not only minimizes tokenization overhead and computational costs but also delivers the highest prediction accuracy,  leveraging FTS and causal graphs to capture essential temporal relationships effectively. This evidence suggests that strategic pre-processing, such as that used by the CGF-LLM method, is critical for optimizing the scalability and cost-effectiveness of LLM-based forecasting methods.



\begin{table*}[!tb]
\centering
\caption{Comparison of Fuzzy LLM and LLM models, with and without parameter freezing, across datasets. All results are based on one-step-ahead forecasting using the NRMSE (Normalized Root Mean Squared Error) metric. }

\begin{tabular}{|c|ccl|ccl|}
\hline
\multirow{2}{*}{\textbf{Datasets}} & \multicolumn{3}{c|}{\textbf{No Freezing}} & \multicolumn{3}{c|}{\textbf{With Freezing}} \\ \cline{2-7} 
 & \multicolumn{1}{c|}{\textbf{CGF-LLM}} & \multicolumn{1}{c|}{\textbf{CG-LLM}} & \multicolumn{1}{c|}{\textbf{LLM}} & \multicolumn{1}{c|}{\textbf{CGF-LLM}} & \multicolumn{1}{c|}{\textbf{CG-LLM}} & \multicolumn{1}{c|}{\textbf{LLM}} \\ \hline
ECONOMICS & \multicolumn{1}{c|}{0.094 $\pm$ 0.018} & \multicolumn{1}{c|}{0.113 $\pm$ 0.017} & 0.104 $\pm$ 0.018 & \multicolumn{1}{c|}{0.201 $\pm$ 0.032} & \multicolumn{1}{c|}{0.238 $\pm$ 0.029} & 0.222 $\pm$ 0.034 \\ \hline
ENERGY & \multicolumn{1}{c|}{0.066 $\pm$ 0.006} & \multicolumn{1}{c|}{0.083 $\pm$ 0.003} & 0.104 $\pm$ 0.004 & \multicolumn{1}{c|}{0.153 $\pm$ 0.009} & \multicolumn{1}{c|}{0.212 $\pm$ 0.012} & 0.246 $\pm$ 0.004 \\ \hline
IoT & \multicolumn{1}{c|}{0.027 $\pm$ 0.003} & \multicolumn{1}{c|}{0.030 $\pm$ 0.008} & \multicolumn{1}{c|}{0.075 $\pm$ 0.005} & \multicolumn{1}{c|}{0.061 $\pm$ 0.006} & \multicolumn{1}{c|}{0.077 $\pm$ 0.067} & 0.082 $\pm$ 0.004 \\ \hline
CLIMATIC & \multicolumn{1}{c|}{0.029 $\pm$ 0.002} & \multicolumn{1}{c|}{0.063 $\pm$ 0.006} & 0.077 $\pm$  0.006 & \multicolumn{1}{c|}{0.135 $\pm$ 0.015} & \multicolumn{1}{c|}{0.116 $\pm$ 0.011} & 0.125 $\pm$ 0.019 \\ \hline
\end{tabular}

\label{tab:nrmse}
\end{table*}

\begin{table}[]
\centering
\caption{Comparison of Text Size and Token Metrics for CGF-LLM, CG-LLM, and LLM Across Datasets}
\label{tab:tokens}
\resizebox{8,6cm}{!}{%
\footnotesize
\begin{tabular}{|c|r|r|r|r|}
\hline
\textbf{Dataset} & \textbf{Metric} & \textbf{CGF-LLM} & \textbf{CG-LLM} & \textbf{LLM} \\ \hline

\multirow{6}{*}{ECONOMICS} 
& Total Text Size & 43,875 & 40,643 & 1,335,281 \\ \cline{2-5} 
& Train Text Size & 35,034 & 32,627 & 1,069,999 \\ \cline{2-5} 
& Test Text Size & 8,841 & 8,016 & 265,282 \\ \cline{2-5} 
& Total Tokens & 17,340 & 32,984 & 846,300 \\ \cline{2-5} 
& Train Tokens & 13,880 & 26,410 & 677,625 \\ \cline{2-5} 
& Test Tokens & 3,460 & 6,574 & 168,675 \\ \hline

\multirow{6}{*}{ENERGY} 
& Total Text Size & 1,276,952 & 1,190,523 & 23,194,823 \\ \cline{2-5} 
& Train Text Size & 1,021,690 & 952,364 & 18,564,729 \\ \cline{2-5} 
& Test Text Size & 255,262 & 238,159 & 4,630,094 \\ \cline{2-5} 
& Total Tokens & 380,451 & 839,680 & 13,434,880 \\ \cline{2-5} 
& Train Tokens & 304,384 & 671,744 & 10,747,904 \\ \cline{2-5} 
& Test Tokens & 76,067 & 167,936 & 2,686,976 \\ \hline

\multirow{6}{*}{IoT} 
& Total Text Size & 3,147,800 & 2,064,031 & 94,793,482 \\ \cline{2-5} 
& Train Text Size & 2,518,273 & 1,651,208 & 75,825,454 \\ \cline{2-5} 
& Test Text Size & 629,527 & 412,823 & 18,968,028 \\ \cline{2-5} 
& Total Tokens & 839,412 & 1,738,840 & 30,699,520 \\ \cline{2-5} 
& Train Tokens & 671,552 & 810,007 & 24,559,616 \\ \cline{2-5} 
& Test Tokens & 167,860 & 347,768 & 6,139,904 \\ \hline

\multirow{6}{*}{CLIMATIC} 
& Total Text Size & 786,258 & 660,437 & 29,934,389 \\ \cline{2-5} 
& Train Text Size & 629,000 & 528,553 & 23,949,302 \\ \cline{2-5} 
& Test Text Size & 157,258 & 131,884 & 5,985,087 \\ \cline{2-5} 
& Total Tokens & 461,076 & 576,400 & 10,731,520 \\ \cline{2-5} 
& Train Tokens & 368,896 & 461,120 & 8,585,216 \\ \cline{2-5} 
& Test Tokens & 92,180 & 115,280 & 2,146,304 \\ \hline

\end{tabular}
}
\end{table}

\section{Conclusion }
\label{sec_conclusao}
This research introduces an LLM-based TSF method named CGF-LLM, which combines the concepts of causal graphs, FTS, and LLMs (fine-tuned GPT-2), applied to MISO applications. The key innovation of this technique relies on converting numerical time series into fuzzy causal text as input for the GPT-2 model, achieved through the parallel use of fuzzification and causality. The generated text provides an interpretable representation of the complex dynamics within the original time series. This method was tested across four different datasets, and the obtained results highlight the efficacy of the proposed CGF-LLM in terms of accuracy and computational costs compared to CG-LLM and regular LLM. 

Despite the success of CGF-LLM, the current study utilized predefined values of hyperparameters. Thus, one possible future direction would be to optimize the model’s hyperparameters via auto ML or other advanced optimization techniques. Another research avenue is to explore other alternatives such as GPT-3, GPT-4, and others. Also, future work may extend CGF-LLM to support multiple-output forecasting tasks.

\section*{Acknowledgment}

This work has been supported by the Brazilian agencies (i) National Council for Scientific and Technological Development (CNPq), Grant no. 304856/2025-8, ``\textit{Aprendizado de Máquina Colaborativo e Proteção à Privacidade}''; (ii) Coordination for the Improvement of Higher Education Personnel (CAPES) through the Academic Excellence Program (PROEX).

Partially funded by the R\&D\&I agreement between UFMG and Fundep -- Research Development Foundation: ``Federated Machine Learning Program for Connected Vehicles''.

The authors acknowledge partial support from Kunumi and Embrapii, project PDCC-2412.0030.



\end{document}